\documentclass[letterpaper, 10 pt, conference]{ieeeconf}  

\IEEEoverridecommandlockouts                              

\usepackage{multirow}
\usepackage{amssymb}

\usepackage[utf8]{inputenc}
\usepackage{color}
\usepackage{graphicx}
\usepackage{makecell}
\usepackage{url}

\date{June 2019}

\begin{document}

\title{\LARGE \bf Topological Descriptors for Parkinson's Disease Classification and Regression Analysis}

\author{Afra Nawar\textsuperscript{1,2}, Farhan Rahman\textsuperscript{1,2}, Narayanan Krishnamurthi\textsuperscript{3}, Anirudh Som\textsuperscript{1,4}, Pavan Turaga\textsuperscript{1,4}}







\maketitle

\begin{abstract}
     
     \textit {At present, the vast majority of human subjects with neurological disease are still diagnosed through in-person assessments and qualitative analysis of patient data. In this paper, we propose to use Topological Data Analysis (TDA) together with machine learning tools to automate the process of Parkinson's disease classification and severity assessment. An automated, stable, and accurate method to evaluate Parkinson's would be significant in streamlining diagnoses of patients and providing families more time for corrective measures. We propose a methodology which incorporates TDA into analyzing Parkinson's disease postural shifts data through the representation of persistence images. Studying the topology of a system has proven to be invariant to small changes in data and has been shown to perform well in discrimination tasks. The contributions of the paper are twofold. We propose a method to 1) classify healthy patients from those afflicted by disease and 2) diagnose the severity of disease. We explore the use of the proposed method in an application involving a Parkinson's disease dataset comprised of healthy-elderly, healthy-young and Parkinson's disease patients. Our code is available at \url{https://github.com/itsmeafra/Sublevel-Set-TDA}.}
     
\end{abstract}

\let\thefootnote\relax\footnotetext{ \\
\textsuperscript{1}School of Electrical, Computer and Energy Engineering, Arizona State University\\ \textsuperscript{2}School of Computing, Informatics, and Decision Systems Engineering, Arizona State University\\\textsuperscript{3} Edson College of Nursing and Health Innovation, Arizona State University\\\textsuperscript{4}School of Arts, Media and Engineering, Arizona State University\\
\indent This work was supported in part by NSF CAREER grant number 1452163 along with an REU supplement and by the National Institute of Child Health and Human Development, NIH 1R21HD060315. ASU’s institutional review board approved all study materials and procedures (protocol number 0808003170).}


\section{Introduction}\label{section_intro}

%

%

Parkinson's disease is currently the second most common neurodegenerative disorder, second only to Alzheimer's, and its incidence is projected to increase with time \cite{reeve2014ageing}. The disease is characterized by the dysfunction of dopamine producing neurons of the substantia nigra in the central nervous system. This degradation ultimately induces chronic, progressive symptoms in patients, predominantly affecting their motor control \cite{demaagd2015parkinson}. Its affected population is staggeringly high --- it is approximated that around 1 million people suffer from the disease in the United States and 10 million are afflicted worldwide \cite{marras2018prevalence,national2009parkinson}. Parkinson's is more prevalent in the elderly than the young, with most patients reported to be over the age of 50 \cite{van2003incidence}. Given their elderly age, these patients may find Parkinson's disease particularly debilitating as they may already be more susceptible to other diseases and the natural effects of aging. Parkinson's patients are also substantially more likely to be hospitalized than the general population \cite{gerlach2011clinical}. 
The most common physical symptoms of Parkinson's disease are muscle rigidity, lack of balance and postural control, tremors, and lack of fine motor skills \cite{national2009parkinson}. Although these symptoms are apparent in almost all patients, detecting Parkinson's disease before symptoms become severe or when they are below the threshold of human perception remains a difficult task.

Parkinson's disease is diagnosed primarily by trained medical professionals using in-person assessments and surveys that gauge Parkinson's disease patients' motor abilities. A common measure of the disease severity is the Unified Parkinson's Disease Rating Scale (UPDRS) score used by clinicians to track symptom progression \cite{ramaker2002systematic}. Because the disease is mainly analyzed by medical personnel, diagnoses are prone to subjectivity. To mitigate this, many have tried to develop machine learning (ML) methods to assess the severity of Parkinson's disease. Some of these automated methods utilize data from different sensor devices. Sensors provide biometric data about a Parkinson's Disease patient which can then be processed to determine discrepancies between patients and healthy subjects. 


Sensors also offer Parkinson's disease patients a more convenient alternative to conventional assessment methods. Patients are no longer required to travel to clinics for diagnoses and checkups or sit through scans which may be difficult for them. Some studies utilize sensor data such as those acquired from inertial measurement units, wrist sensors, accelerometers, gyroscopes and voice recordings to name a few \cite{eskofier2016recent,kim2018wrist,kubota2016machine}. Our proposed method utilizes pressure platform data to maximize convenience by allowing patients to set up an in-house automated assessment system \cite{krishnamurthi2012deep}. We aim to show that by using topological descriptors of a patient's postural shift data, we can find an effective and efficient means for Parkinson's disease analysis. Particularly, we use tools from Topological Data Analysis (TDA) to classify Parkinson's disease patients from healthy subjects, and predict the severity to which a patient may be affected.

    

The focus on TDA techniques is inspired by previous successes in time-series and systems analysis \cite{gholizadeh2018short}. Currently, TDA methods have shown promise for classification in a variety of tasks including 3D shape retrieval, hand gesture recognition, and texture classification 
\cite{li2014persistence}. Such results allow us to be optimistic that an application in disease analysis should produce significant results.



\textbf{Paper Organization:} In Section \ref{section_related_work}, we touch on related work on similar problems and previously used techniques. Section \ref{section_background_concepts} provides background information on the techniques used in our experiments. Section \ref{section_materials_and_methods} outlines the method that we propose using for the classification and regression problems on the data of interest. Section \ref{section_experimental_results} provides a summary of the results of our experiments. In Section \ref{section_conclusion_future_work}, we discuss the implications of our results and propose ideas for future work.


\section{Related Work}\label{section_related_work}


%

%


The last decade has seen a vast improvement in the efficiency and computation time of machine learning algorithms \cite{sejnowski2018deep}. Prior to this, although ML concepts were well developed, major bottlenecks to their widespread use included availability of data and speed of computation. Because of its promise in many areas of classification and regression analysis, many have approached the task of Parkinson's disease assessment from an ML standpoint. As Parkinson's disease progression is marked by a myriad of effects, many measures of aptitude have been used in these studies. 

Eskofier \textit{et al.} used inertial measurement units to collect patient data and utilized deep learning to detect the presence of the Parkinson's disease symptom, bradykenesia \cite{eskofier2016recent}. Kim \textit{et al.} collect patient data through wrist sensors of a custom-developed wearable to detect tremor severity \cite{kim2018wrist}. Kubota \textit{et al.} summarizes multiple authors' work on using machine learning techniques with large-scale wearable sensors \cite{kubota2016machine}.


Although these methods have shown substantial promise in becoming efficient tools for Parkinson's disease diagnosis, machine learning techniques are not invariant to small changes in input data. In order to make machine learning techniques robust to such changes, data augmentation must be performed and care must be taken to cover every edge case. This process is time consuming and is not bulletproof - it is difficult to assert that every scenario has been accounted for. TDA can help mitigate these effects as it is invariant to smooth deformations (stretching and bending but not tearing or gluing) \cite{som2018perturbation}, \emph{i.e.}, changes of the data-space including rotations, translations, scaling, and small perturbations negligibly change the computed topological representation \cite{som2018perturbation}. This is especially significant in medical datasets because we can account for slight differences between patient data measurements taken in different conditions. Our inspiration for this paper comes from the work of Som \textit{et al.} which employs shape distributions for Parkinson's disease analysis \cite{som2017multiscale,som2016attractor}. For further reading, \cite{somgeometric} surveys recent topological representations and their associated metrics.


Shape distributions and TDA both provide unique advantages to analyzing dynamic postural shifts data. For the Parkinson's disease dataset referenced later in this paper, and used by \cite{som2017multiscale,som2016attractor}, both TDA and shape distributions allow for topological representations to be obtained from variable length time-series signals. These representations are also invariant to the order in which the target reach tasks were performed within each trial of the dataset. However, although shape distributions have proven significant in classification problems, TDA is a more simple and mathematically grounded technique comparatively. This is because it provides stable results using fewer intermediate steps, \emph{i.e.}, embedding delay, embedding dimension, and histogram bins used in shape distributions are not required in TDA. 
 
When using TDA, the most commonly used descriptor is the persistence diagram. Persistence diagrams provide a simple yet robust way to encode time-series data. They are provably stable, summarize topology of the underlying data in a compact form, and are effective in discriminating between configurations of functions that lie on different domains \cite{chazal2009proximity,cohen2007stability}. A major obstacle to their use is that they cannot be directly incorporated into ML algorithms as most algorithms expect feature-vector representations of the data in the $\mathbb{R}^n$ space. It is also computationally expensive to compare persistence diagrams, with the computational complexity in the order of $\mathcal{O}(n^3)$, where $n$ is number of points in each persistence diagram. 


This problem can be solved by using persistence images -- a vectorized, yet stable, form of the persistence diagram \cite{adams2017persistence}. Persistence images encode a persistence diagram in vectorized form by fitting a weighted Gaussian over each point on the diagram. Using persistence images has shown to perform comparably to persistence diagrams in terms of classification accuracy and require far less computation time to compare persistence images \cite{adams2017persistence}. They also perform significantly better when compared to other existing representations of persistence diagrams such as persistence landscapes \cite{adams2017persistence}. 

As many datasets can be described as a noisy sampling of the underlying space, TDA can be used to capture its properties \cite{chazal2009proximity}. TDA has been shown to be useful in datasets with large numbers of discrete points \cite{gholizadeh2018short}, which may be useful in large-scale applications involving biomechanical data. It has also been shown to provide results in applications involving stability determination, and in the analysis of periodic behavior of signals and systems \cite{gholizadeh2018short}. As human body movements are quasi-periodic and have been theorized to follow a dynamical system, TDA shows promise in becoming an effective and stable means for postural analysis 
\cite{van2000variability}. We propose to use persistence images to model Parkinson's disease patient dynamic shift data as an effective and promising way to diagnose the disease and assess its severity due to its robustness and generalizability.

\begin{figure}[t]
	\centering
	\vspace{0.075in}
	\includegraphics[width=1\linewidth]{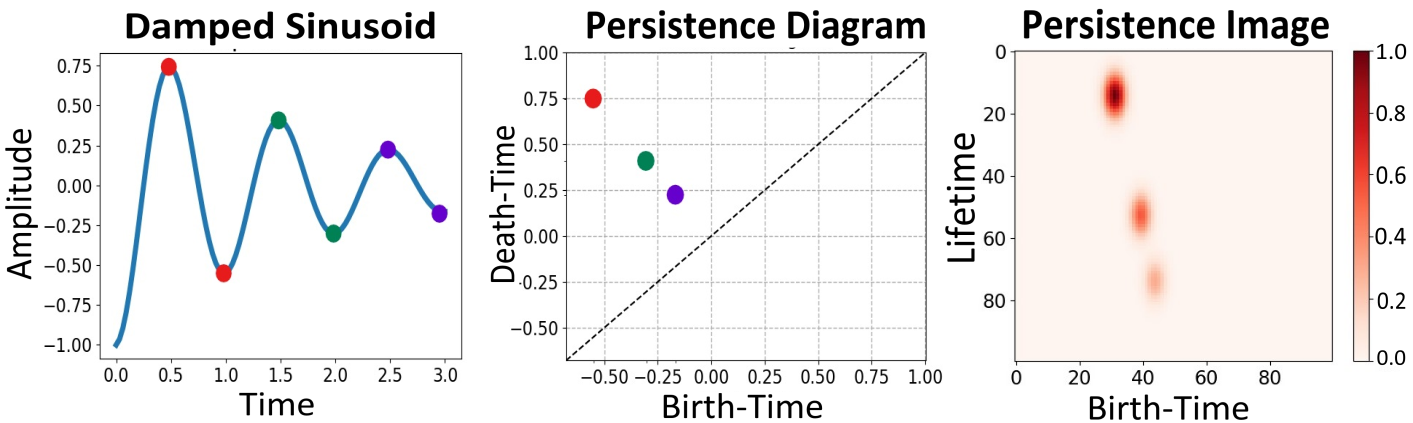}
	\vspace{-0.25in}
	\caption{Example time series signal paired with its corresponding persistence diagram and persistence image. Here we use a damped sinusoid as the original signal. The \emph{x, y} coordinates in the persistence image correspond to grid values.}\label{signal_pd_pi}
	\vspace{-0.225in}
\end{figure}

\section{Background}\label{section_background_concepts}



%

\begin{figure*}[t]
	\centering
	\vspace{0.075in}
	\includegraphics[width=0.925\linewidth]{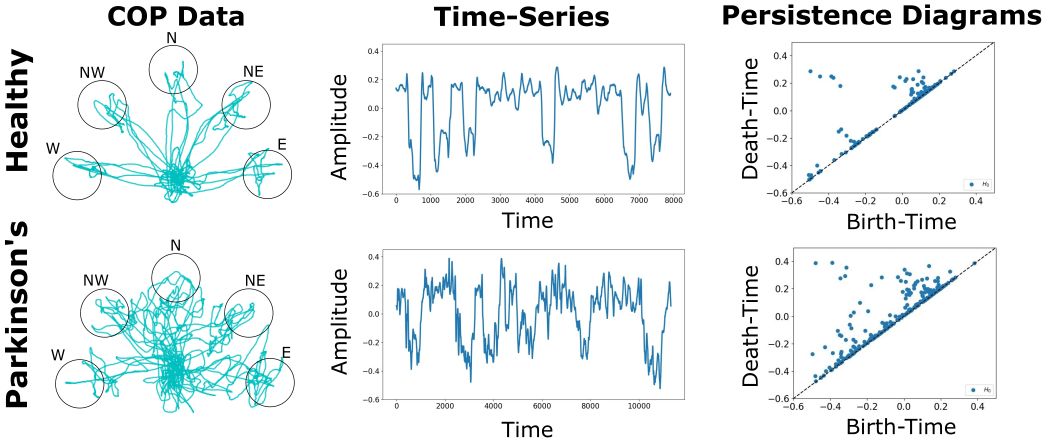}
	\vspace{-0.1in}
	\caption{Illustration of the 2D dynamic postural shifts data (left column), its corresponding \emph{x}-coordinate signal (middle column) and persistence diagram (right column) for a healthy and Parkinson's disease subject. Target reach task during the dynamic postural shifts protocol did not follow a particular order.}\label{pd_hs_illustration}
	\vspace{-0.2in}
\end{figure*}

Before discussing the method with which we classify Parkinson's disease patients, we wish to provide the reader with some background information on the topological descriptors used to capture each patient's data: persistence diagrams and persistence images. For a more detailed explanation of these descriptors, refer to \cite{edelsbrunner2000topological,edelsbrunner2010computational,adams2017persistence}.

\noindent \textbf{Topological Data Analysis and Persistence Diagrams:}
For a graph $\mathcal{G = \{V, E\}}$, we can define $\mathcal{V}$ to be a set of vertices and $\mathcal{E}$ to be the set of edge connections between vertices in $\mathcal{G}$. Using scalar field topology, we can define a function $f: \mathcal{V} \to \mathbb{R}$ on the vertices of a graph $\mathcal{G}$. With this function defined, we can then use the union of the sub-level sets $\mathbb{F}^\alpha = f^{-1}([-\infty,\alpha])$ to represent the mapping process of peaks to troughs on a persistence diagram. We define the birth-time $b$ of a trough at $u$ as the value of $f(u)$, where $f(v) > f(u)$ for every $v$, such that $(u,v) \in \mathcal{E}$.  As $\alpha$ is increased from $-\infty$ to $\infty$, we define the connected sub-graph in $\mathbb{F}^\alpha$ that includes $u$ as $C(u,\alpha)$. We define the death-time $d$ of trough $u$ as the smallest value of $\alpha > f(u)$ such that $u$ is the global minimum of $C(u,\alpha)$. Additionally, the lifetime $l$ of a peak $b$ is defined by the absolute difference between its death-time and birth-time, \emph{i.e.}, $l = |d - b|$. A point on the persistence diagram is said to be more \textit{prominent} if its lifetime is greater. This means that the persistence diagram point is further away from the line defined by $y=x$ where $y$ is the death-time and $x$ is the birth-time. A persistence diagram can then be created as a multi-set of points $(b, d)$. Figure \ref{signal_pd_pi} illustrates a damped sinusoid signal and its corresponding persistence diagram. Each trough (birth) is mapped to a peak (death) and is color coded for convenience. These pairs (3 total) are then mapped on the persistence diagram. The amplitude of the peak and the trough correspond to the death-time and birth-time respectively, which forms a point given by a coordinate pair on the persistence diagram. We would like to note that the filtration process followed above falls under a field referred to as \textit{scalar} field topology as each vertex of the persistence diagram is mapped to a real number. Another way to create persistence diagrams is through a process known as \textit{persistent homology} which captures the graphs shape by summarizing the $k$-dimensional holes in the data. More information on this process can be found in \cite{edelsbrunner2000topological}.

\noindent \textbf{Persistence Images:}
Persistent diagrams are a good representative of the shape of a time-series signal, but are difficult to analyze through conventional machine learning algorithms. Persistence diagrams also assign the same significance to every point. However, in certain applications, it may be desirable to regard some points in a persistence diagram as more significant than others. In order to conserve the information contained in a persistence diagram while solving these issues, we can convert the persistence diagram to a persistence image by performing the following. Given a persistence diagram $B$, we first apply a linear transformation $T: \mathbb{R}^2 \to \mathbb{R}^2$ to rotate the persistence diagram, where $T(x,y) = (x,y-x)$ with $x$ referring to birth-times and $y$ referring to death-times. We then map the persistence diagram to a continuous, integrable weighting function $p: \mathbb{R}^2 \to \mathbb{R}$ referred to as a persistence surface. This persistence surface is a weighted sum of Gaussians centered on each point in the persistence diagram. We can then divide the persistence surface into a discretized grid. An integration is then performed over each grid section, and each grid point is assigned a scalar value, giving us a vectorized, weighted image to represent the persistence diagram. This persistence image can then be used in traditional machine learning algorithms, and the weighting function and grid resolution can be tuned for better performance based on the application. Though other weighting schemes can be chosen, the most commonly used weighting function assigns a higher significance to points with a higher lifetime.

Figure \ref{signal_pd_pi} shows the conversion of a persistence diagram to a persistence image. During conversion, the point with the largest lifetime, corresponding to the most prominent peak and trough is the brightest point, highlighting its significance. In this way, noisy data can still be accurately represented by a persistence image as smaller perturbations will likely be less visible and thus less significant.

\section{Extracting Persistence Images from Time-Series Signals}\label{section_materials_and_methods}

%

In this section we highlight the steps performed to generate persistence images from a dynamic postural shifts dataset. Detailed description of the dataset and the protocol used to collect the data can be found here \cite{krishnamurthi2012deep}. We briefly describe the dataset below for the reader's convenience.

The dataset consists of dynamic postural shifts data from a total of 60 subjects from three different classes -- 21 healthy-young, 22 healthy-elderly, and 17 Parkinson's disease subjects. Each healthy subject performed 5 trials of the protocol, while each Parkinson's disease patient performed 3 trials, giving a total of 266 trials. Each trial contains data describing time series information of the subject's center-of-pressure (CoP). CoP data was collected in the medio-lateral and antero-posterior direction (referred to as $x$ and $y$ respectively in this paper), as well as its forces and moments in 3D cartesian space ($f_x$, $f_y$, $f_z$ and $m_x$, $m_y$, $m_z$ respectively). Figure \ref{pd_hs_illustration} illustrates an example of the 2D dynamic postural shifts data for a healthy and Parkinson's disease subject. In the same figure we also illustrate the \emph{x}-coordinate signal of the dynamic shifts task and its corresponding persistence diagram.

The task of generating persistence images from the time-series data gathered from these subjects is split into two steps: 1) preprocessing the data so that it is in a form that is easily comparable between subjects and 2) computing the persistence images from the data.

\noindent \textbf{Data Preprocessing:} In order to accurately compare the data from each subject to another, a series of normalization steps were taken prior to computing the persistence images. The time series signals from each segment of data for each trial ($x,y,f_x,f_y,f_z,m_x,m_y,m_z$) were first zero-normalized -- i.e. the data was centered around the x-axis. To do this, the mean of each time-series signal was found and subtracted from every data point. After the data was zero-normalized, the range of the data was enforced to be $[-1$, $1]$. This was done by finding the maximal magnitude ranges 
from all trials across all subjects for each segment of data. After this, each trial's zero-normalized data points were divided by their respective maximal magnitude range. This ensured that the time-series data points of all subjects (a) were in the same range and (b) were scaled evenly.  

\noindent \textbf{Generating Persistence Diagrams and Persistence Images:} To compute persistence diagrams from the normalized time series signals, we used the Ripser Python package in Scikit-TDA \cite{scikittda2019}. After computation, we thresholded the persistence diagrams to reduce points that may not be significant. To do this we first computed the lifetime of every point on the persistence diagram. We then chose a thresholding value, $t$. Any point with \emph{lifetime} $< t$ was removed from the persistence diagram. This change is analogous to removing points that have low persistence and can therefore be considered as noisy components in the signal. 

To compute the persistence images, we used the Python package Persim in Scikit-TDA \cite{scikittda2019}. The modifiable parameters for computing persistence images included grid size, Gaussian kernel standard deviation, and birth-time range. We experimented with different combinations of each parameter, however, our best results came from the setting: 50x50 grid size, 0.03 standard deviation, $[-0.5$, $0.5]$ birth-time range for $f_x$, $[-0.75$, $0.75]$ for $m_z$ and $[1.5$, $-1.5]$ for all other segments. The discrepancy in birth-time ranges allowed for less-sparse persistence images focused on the range in which their respective persistence diagram's points resided.

\section{Experiments}\label{section_experimental_results}







After creating the persistence images, a complete vectorized descriptor was created for each trial by concatenating its $x, y, f_x, f_y, m_x, m_y,$ and $m_z$ persistence images together in that order. The $f_z$ signal was disregarded due to incompatibilities with computing persistence diagrams for every subject. As each image contains 2500 values (because of the chosen 50x50 grid) the end result of the total descriptor is a vector of size 17500.


The remainder of this section can be split according to our two objectives for analyzing the Parkinson's disease dataset. First, we conduct the classification experiment to categorize the subjects into sub-classes. Second, we do the regression experiment to assess disease severity.

\noindent \textbf {Classification:}
For the classification experiment, we performed two tests on the dataset. The first was a binary classification test to assess whether we can discriminate between a subject with Parkinson's disease and one who is healthy. Our second test, a 3-class classification, is more challenging as we seek to discriminate between healthy-elderly, healthy-young, and Parkinson's disease subjects. For both tests, we used a linear-SVM from the SVM package in the Python library scikit-learn \cite{scikit-learn}. An SVM was chosen because of its computational efficiency for small datasets. To assess the performance of the SVM we performed a round-robin leave one subject out cross-validation in both tests. Out of 60 subjects, we used 59 subjects' trials as our training set and the remaining subject's trials as the test set. In the binary classification experiment, training a SVM using persistence images produced a \textbf{98.87\%} classification accuracy. For the 3-class experiment, we compare our method to baselines such as Peak Velocity Index, Largest Lyapunov Exponent (LLE), and multiple Shape Distribution functions. The peak velocity is defined as the maximum difference computed between adjacent samples in the time-series signal when a subject reaches toward a target (refer to figure 2). The LLE is a measure of the rate of divergence of neighboring trajectories \cite{abarbanel2012analysis}. Shape distributions represent the signature of an object by sampling a predefined function which measures the object's geometric properties \cite{osada2002shape,som2016attractor,som2017multiscale}. The results from the comparison are contained in Table \ref{classification_regression_table}. The classification accuracy of the linear-SVM using persistence images is higher than any of the other measured techniques. Our best results came from training a SVM with regularization parameter equal to 1.5 and using an L1 norm penalty.

\noindent \textbf {Regression:}
For the regression experiment, clinically assigned motor UPDRS scores were used as the ground truth severity for Parkinson's disease patients while all healthy subjects were given a UPDRS score of zero. The same leave-one-subject-out approach was also used on the dataset for the regression experiment. However, a subject's predicted UPDRS score was taken as an average of its trials. Any negative predicted scores from the SVM were fixed to zero. The regression scores achieved from training a SVM using persistence images are comparable to the baseline methods. We achieve a high Pearson-correlation coefficient of 0.8493 and a p-value on the order of 10$^{-18}$. The regression results are also tabulated in Table \ref{classification_regression_table}. Our best results came from training a SVM with regularization parameter equal to 0.85 and using an L1 norm penalty.



\begin{table}[t!]
	\centering
	\vspace{0.1in}
	\scalebox{0.9}{
	\begin{tabular}{ |c|c|c|c|c| } 
			\hline
			\multirow{2}{*}{ \textbf{Method}} & \multirow{2}{*}{\makecell{\textbf{3-Class} \\ \textbf{Classification (\%)}}} & \multicolumn{2}{c|}{\textbf{Regression}} \\ 
			\cline{3-4} &
			& \textbf{Correlation} & \textbf{P-Value} \\
			\hline
			Peak Velocity Index & $53.01$  & $0.8153$ & $2.8227 e^{-15}$ \\
			LLE & $47.37$  & $0.6449$ & $2.6707 e^{-08}$ \\
			A3 \cite{som2016attractor} & $60.53$  & $0.7518$ &  $4.4376 e^{-12}$ \\
			D1 \cite{som2016attractor} & $70.30$  & $0.8479$ &  $9.2763 e^{-18}$ \\
			D2 \cite{som2016attractor} & $71.43$  & $0.9006$ &  $1.1847 e^{-22}$ \\
			D3 \cite{som2016attractor} & $68.42$  & $0.8479$ &  $1.2715 e^{-17}$ \\
			D4 \cite{som2016attractor} & $65.79$  & $0.8509$ &  $7.2852 e^{-18}$ \\
			A3M \cite{som2017multiscale} & $65.41$  & $0.8528$ &  $5.2806 e^{-18}$ \\
			D1M \cite{som2017multiscale} & $73.31$  & $0.8649$ &  $5.1880 e^{-19}$ \\
			D2M \cite{som2017multiscale} & $71.05$  & $0.8639$ &  $6.3778 e^{-19}$ \\
			D3M \cite{som2017multiscale} & $71.80$ & $0.8598$ &  $1.4184 e^{-18}$ \\
			D4M \cite{som2017multiscale} & $68.05$ & $0.8680$ &  $2.7529 e^{-19}$ \\
			\hline
			\textbf{Persistence Images} & $\mathbf{73.68}$ & $\mathbf{0.8493}$ & $\mathbf{9.85 e^{-18}}$ \\
			\hline
	\end{tabular}}
	\caption{3-class results for Persistence Image Classification and Regression tests performed on the postural shifts dataset. }\label{classification_regression_table}
	\vspace{-0.225in}
\end{table}

\section{Conclusion and Future Work}\label{section_conclusion_future_work}



In this paper, TDA, particularly that of extracting persistence images from time-series data has been used to improve the ability to classify Parkinson's disease patients and assess disease severity. 
Our methods show promise toward providing more accurate classifications than previously used baselines in a stable, robust, and efficient way and could thus be significant in future medical applications. 

In the future, we plan to extend our analysis to other Parkinson's disease datasets (e.g. those with different sensors or data collection protocols). We also plan to determine if there are other ML techniques that when paired with persistence images would yield better results than those referenced in this paper. We are particularly interested in unsupervised learning and deep learning techniques. We would also like to explore the use of persistence images in neurodegenerative or movement disorders other than Parkinson's to assess whether its usage can be significant in these spaces as well.  



{\small
\bibliographystyle{ieee}
\bibliography{egbib}
}
\end{document}